\documentclass[preprint]{elsarticle}

\usepackage{graphicx}
\usepackage{url}
\usepackage{latexsym}
\usepackage{amsthm}

\newtheorem{theorem}{Theorem}
\newtheorem{example}{Example}
\newtheorem{corollary}{Corollary}

\begin{document}

\title{Measuring Similarity of Graphs and their Nodes by Neighbor Matching}

\author{Mladen Nikoli\'c\corref{cor1}}
\ead{nikolic@matf.bg.ac.rs}
\address{Faculty of Mathematics, University of Belgrade, Studentski Trg 16, Belgrade, Serbia}
\cortext[cor1]{Faculty of Mathematics, University of Belgrade, Studentski Trg 16, Belgrade, Serbia}

\date{}

\begin{abstract}
The problem of measuring similarity of graphs and their nodes
is important in a range of practical problems.
There is a number of proposed measures, some of them being based on iterative
calculation of similarity between two graphs and the principle
that two nodes are as similar as their neighbors are. In our work,
we propose one novel method of that sort, with a refined concept
of similarity of two nodes that involves matching of their neighbors.
We prove convergence of the proposed method and show that it has some
additional desirable properties that, to our knowledge, the existing methods lack.
We illustrate the method on two specific problems and empirically compare
it to other methods.
\end{abstract}


\maketitle

\section{Introduction}
\label{sec:introduction}

Many or most data analysis techniques are designed for data that are
represented by vectors of numbers.
However, this kind of representation often leads to loss of structural
information contained in the original data, while preserving structural
information may be essential in some applications.
This requires a richer problem representation and corresponding
data analysis techniques. For example, in many practical domains,
structural information in the data can be represented using graphs.

Similarity measures between objects are of central importance for
various data analysis techniques. The same holds for the special
case of similarity between graphs and a number of measures for
this purpose have been proposed.
In this paper, we focus on iterative methods relying on the principle
that the nodes of two graphs are as similar as their neighbors in
respective graphs are \cite{Kleinberg99, Heymans03, Blondel04,Zager08}.
These methods have been successfully applied in several domains like
adequate ranking of query results \cite{Kleinberg99}, synonym extraction \cite{Blondel04}, database
structure matching \cite{Melnik02}, construction of phylogenetic trees \cite{Heymans03},
analysis of social networks \cite{Leicht06}, etc.

In this paper, we try to identify desirable properties not present in the existing methods for
measuring similarities of graph nodes. We propose a refinement
of the notion of similarity of two nodes which leads to a new
method for measuring similarities of graph nodes and similarities
of graphs. We prove convergence of the proposed method and show
that it has some additional desirable properties that, to our knowledge, the existing methods
lack.

We implemented the proposed method and evaluated it on two problems in order to illustrate
that our method can capture the notion of similarity useful in practical problems.
The first test problem was finding a subgraph of a graph that is isomorphic
to some other given graph. The second test problem was the classification of Boolean
formulae based on their underlying graph structure.

The rest of the paper is organized as follows.
In Section \ref{sec:preliminaries}, we present the preliminaries used in this
paper. Existing methods are described and analyzed in Section \ref{sec:relatedwork}.
In Section \ref{sec:method} we present our new method --- the method of neighbor
matching and prove its properties. Results of experimental evaluation
and comparison to other methods are given
in Section \ref{sec:evaluation}. In Section \ref{sec:conclusions}, we draw final
conclusions and give some directions of the future work.

\section{Preliminaries}
\label{sec:preliminaries}

A directed graph $G=(V,E)$ is defined by its set of nodes $V$ and its set of edges $E$.
There is an edge between two nodes $i$ and $j$ if $(i,j)\in E$. For the edge
$e=(i,j)$, the {\em source node} is the node $i$, and the {\em terminating node} is the node $j$.
We denote them respectively with $s(e)$ and $t(e)$. We say that the node $i$
is an {\em in-neighbor} of node $j$ and that node $j$ is an {\em out-neighbor} of the
node $i$ if $(i,j)\in E$. An {\em in-degree} $id(i)$ of the node $i$ is the number of
in-neighbors of $i$, and an {\em out-degree} $od(i)$ of the node $i$ is
the number of out-neighbors of $i$.
A {\em degree} $d(i)$ of the node $i$ is the sum of in-degree and out-degree of $i$.
Two graphs are \emph{isomorphic} if there exists a bijection $f:V_A\rightarrow V_B$,
such that $(i,j)\in E_A$ if and only if $(f(i),f(j))\in E_B$. An isomorphism of a graph
$G$ to itself is called \emph{automorphism}. A \emph{colored graph} is a graph
in which each node is assigned a color. For colored graphs, the definition of
isomorphism additionally requests that nodes $i$ and $f(i)$ have the same color.
A \emph{random Erd\H{o}s--R\'enyi graph} $G_{n,p}$ is a graph with $n$ nodes in
which each two nodes share an edge with probability $p$ \cite{Erdos59}.
A graph $G_B$ is an \emph{induced subgraph} of a graph $G_A$ if $V_B\subseteq V_A$ and
for each pair of nodes $i,j\in V_B$ it holds $(i,j)\in E_B$ if and only if
$(i,j)\in E_A$.

The \emph{similarity measure} $s$ is a function $s:D_1\times D_2\rightarrow R$ where
$D_1$ and $D_2$ are possibly equal sets of objects. A higher value of similarity
measure should imply a higher similarity in some intuitive sense. Choice of a
similarity measure to be used in some context is often guided by its usefulness
in practice.

Similarity measure over the nodes of two graphs can be represented
by a \emph{similarity matrix} $X=[x_{ij}]$ of dimension $|V_A|\times|V_B|$
with the element $x_{ij}$ denoting a similarity of the nodes $i\in V_A$ and
$j\in V_B$.

Let $A$ and $B$ be two finite sets of arbitrary elements.
A \emph{matching} of elements of sets $A$ and $B$ is a set of pairs $M=\{(i,j)|i\in A,j\in B\}$
such that no element of one set is paired with more than one element of the other set.
For the matching $M$ we define \emph{enumeration functions} $f:\{1,2,\ldots k\}\rightarrow A$
and $g:\{1,2,\ldots k\}\rightarrow B$ such that $M=\{(f(l),g(l))|l=1,2,\ldots,k\}$ where $k=|M|$.
Let $w(a,b)$ be a function assigning weights to pairs of elements $a\in A$ and $b\in B$.
The goal of the \emph{assignment problem} is to find a matching of elements of $A$ and $B$
with the highest sum of weights (if two sets are of different cardinalities, some elements of the larger
set will not have corresponding elements in the smaller set).
The assignment problem is usually solved by the well-known Hungarian
algorithm of complexity $O(mn^2)$ where $m=\max(|A|,|B|)$ and $n=\min(|A|,|B|)$ \cite{Kuhn55}.
There are more efficient algorithms, such as one due to Edmonds and Karp of complexity
$O(mn\log n)$ \cite{Edmonds72} and even
more efficient one, due to Fredman and Tarjan of complexity $O(mn+n^2\log n)$ \cite{Fredman87}.

\section{Existing Methods for Measuring Graph Node Similarity}
\label{sec:relatedwork}

In this section we briefly describe relevant iterative methods for
measuring similarity of graph nodes and we try to identify some
desirable properties that they lack.

Assume that two directed graphs
$G_A=(V_A,E_A)$ and $G_B=(V_B,E_B)$ are given. Iterative methods
calculate similarity of nodes of these two graphs by repeatedly
refining the initial estimate of similarity using some update
rule of form $[x^{k+1}_{ij}]\leftarrow f([x^k_{ij}])$.
Iterations are performed until some termination condition
is met. At the end, the similarity matrix $X=[x_{ij}]$ is produced.
Different rules for update of similarity of two nodes are proposed.
They usually include summing all the similarities between the
neighbors of first node and the neighbors of the second node.

One of the first influential iterative approaches is due to
Kleinberg \cite{Kleinberg99}, further generalized by Blondel et al.~\cite{Blondel04}.
In the method of Blondel et al. the update rule for $x_{ij}$ in step $k+1$ is given by
$$x^{k+1}_{ij}\leftarrow \sum_{(p,i)\in E_A,(q,j)\in E_B}x^{k}_{pq}+\sum_{(i,p)\in E_A,(j,q)\in E_B}x^{k}_{pq}.$$
The similarity matrix $X$ is normalized by $X\leftarrow X/\|X\|_2$ after each step.

The earlier approach by Melnik et al.~\cite{Melnik02} can be seen as a more general version of
of this method where the similarities between neighbor nodes $x^{k}_{pq}$ are weighted.

The method of Blondel et al.~was modified by Zager and Verghese \cite{Zager08} to take into account
similarity of the edges too. The update rule for the edge similarity matrix $Y=[y_{uv}]$, where
$u\in E_A$ and $v\in E_B$, is given by
$$y^{k+1}_{uv}\leftarrow x^{k}_{s(u)s(v)}+x^k_{t(u)t(v)}.$$
The update rule for similarity of nodes is then given in terms of similarities of the edges
$$x^{k+1}_{ij}\leftarrow \sum_{t(u)=i,t(v)=j}y^{k}_{uv}+\sum_{s(u)=i,s(v)=j}y^{k}_{uv}.$$
Matrix normalization of the similarity scores is applied in this approach too.

The approach by Heymans and Singh \cite{Heymans03} is somewhat different
and more complex than the described methods, and we
only briefly mention its most important aspects. In order to estimate similarity
in each iteration, similarity terms and dissimilarity terms are calculated,
based on the similarity scores of the previous iteration. These terms average the similarities
of the in-neighbor and similarities of the out-neighbors. Similarity terms are calculated both for
the original graphs and their complements. Dissimilarity terms are
calculated using one graph and the complement of the other, and vice versa.
Dissimilarity terms are subtracted from similarity terms to obtain new estimate of similarity scores.
The matrix normalization is performed after each iteration.

There are approaches that are designed for measuring similarity between
the nodes of the same graph \cite{Jeh02,Leicht06}. We don't discuss these
methods as they are less general than the former ones.

The described methods lack some desirable and natural properties.
Of course, not all the method lack all the listed properties.

\paragraph{If the graphs is compared to itself, each node should be most similar to itself}
This is a natural property, expected for all similarity measures.
Nevertheless, for all mentioned methods it is easy to construct graphs
for which there is a node which is more similar to some other node of the same graph than to itself.
This can easily occur, for instance, in methods where the update rule consists of simple summation of similarities of
neighbor nodes. This results in nodes of higher degree having more terms in the summation
and hence, higher similarity with other nodes \cite{Cason10}.

\paragraph{Similarity scores should have a fixed range with similarity of a node to itself always
taking the maximal value} It is customary for similarity measure in general (not only for
similarity measures for graphs) to have a fixed range (e.g., from 0 to 1 or from -1 to 1).
Without the loss of generality, we will assume the range $[0,1]$.
Also, similarity of each object to itself should be 1.
These properties facilitate intuitive understanding of similarity scores.
Well-known examples of measures for which these requirements are
fulfilled are cosine, correlation coefficient, Jaccard coefficient, etc.
However, the mentioned methods for calculating graph node similarity lack
this property. When the similarity scores are calculated
for the nodes of the same graph, the similarity score of one node compared to itself
can be different from the similarity score of some other node compared to itself.
So, one node can be more similar to itself than the other.

It is reasonable to make even stricter requirement: if two graphs $G_A$ and $G_B$ are
isomorphic, with isomorphism $f:V_A\rightarrow V_B$, the similarity score $x_{if(i)}$ should be 1 for
all $i\in V_A$.

\paragraph{Similarity scores should be meaningful in absolute terms}
Due to the normalization of the similarity matrix, one similarity score
$x_{ij}$ can change only if other similarity scores change accordingly. This makes additional
interdependence between similarity scores that is not a result of the topology of
two graphs alone. It actually means that similarity scores can only reflect
similarity of nodes of two graphs relative to each other. We can't conclude if two nodes are similar,
but only if one pair of nodes is more similar than some other pair of nodes.

Consider the following special case. Suppose that all the nodes of one graph
are equally similar to all the nodes of the second graph. In a normalized matrix
it is impossible that all the similarity scores are equal to 0, or that all the
similarity scores are equal to 1. Because of the normalization constraint, we
can't differentiate between all possible levels of similarity. All we can say
is that the nodes of one graph are equally similar to all the nodes of the
second graph, but not how much.

It would be good if similarity scores don't represent
relative magnitudes of similarities of pairs of nodes,
but in a way ``absolute'' magnitudes with possibility
of all scores having 0 or the maximal value.

The lack of this property, also makes it harder to use similarity scores of the nodes to construct the similarity
measure of whole graphs. Heymans and Singh \cite{Heymans03} were able to achieve this because they use similarity scores
that can be negative (as the consequence of subtracting dissimilarity scores that they use),
but as discussed in the previous special case, it would not be possible with other methods.

\paragraph{If two nodes don't have ingoing or outgoing edges, they should be considered similar}
To our knowledge, this property is present only in the method of Heymans and Singh.
We believe that concepts of in-similarity and out-similarity should be
recognized. Moreover, in-similarity and out-similarity should be 1 if there are
no in-neighbors or out-neighbors.

\section{Method of Neighbor Matching}
\label{sec:method}

In this section we refine the notion of node similarity. Based on that refinement,
we describe a new method (we call this method \emph{the method of neighbor matching})
for measuring  similarity of nodes of graphs and prove its properties.
Then, we define a measure of similarity of whole graphs based on the similarities of
their nodes.

\subsection{Notion of Similarity of Graph Nodes}

In the existing methods, the calculation of similarity $x_{ij}$ is based on adding or
averaging the similarities of all the neighbors of node $i\in V_A$ to all the neighbors of node $j\in V_B$.
We propose a modification to that approach, illustrated by the following intuition.
We perceive our two hands to be very similar, but not because
all the fingers of the left hand are very similar to all the fingers of the right hand, but rather because
of the property that to each finger of the left hand corresponds one finger of the right hand that is very
similar to it. By analogy, the concept of similarity can be refined --- \emph{two nodes $i\in V_A$ and $j\in V_B$
are considered to be similar if neighbor nodes of $i$ can be matched to similar neighbor nodes of $j$}
(hence the name neighbor matching).

\subsection{Measuring Similarity of Graph Nodes}

As in other related methods, similarity scores are calculated as the fixed point of the iterative
procedure defined by some update rule.
In our method, we will differentiate between in-similarity $s_{in}$ and out-similarity $s_{out}$
and will give them equal weights. In order to calculate in-similarity, the
matching of in-neighbors with maximal sum of similarities (as described in Section \ref{sec:preliminaries})
has to be constructed, and analogously for out-similarity.
More formally, the update rule is given by
$$x^{k+1}_{ij} \leftarrow \frac{s^{k+1}_{in}(i,j)+s^{k+1}_{out}(i,j)}{2}.$$
In and out similarities are defined by
\begin{equation}
s^{k+1}_{in}(i,j) \leftarrow \frac{1}{m_{in}}\sum_{l=1}^{n_{in}}x^k_{f^{in}_{ij}(l)g^{in}_{ij}(l)}\hspace{1cm}s^{k+1}_{out}(i,j) \leftarrow \frac{1}{m_{out}}\sum_{l=1}^{n_{out}}x^k_{f^{out}_{ij}(l)g^{out}_{ij}(l)}
\label{eq:updaterule}
\end{equation}
$$m_{in}=\max(id(i),id(j))\hspace{1cm}m_{out}=\max(od(i),od(j))$$
$$n_{in}=\min(id(i),id(j))\hspace{1cm}n_{out}=\min(od(i),od(j))$$
where functions $f^{in}_{ij}$ and $g^{in}_{ij}$ are the enumeration functions of the optimal
matching of in-neighbors of nodes $i$ and $j$ with weight function $w(a,b)=x^k_{ab}$.
In the equation \ref{eq:updaterule}, we define $\frac{0}{0}$ to be 1. This convention ensures
that the similarity of nodes with no in or no out neighbors is recognized. If there is a
difference in the number of in or out neighbors, that difference is penalized when calculating
corresponding similarities since $m_{in}$ and $m_{out}$ are greater than the number of
terms in the summation (which are each less or equal to 1 as we show later).

This method is easily extended to colored graphs. By definition, we can set $x^k_{ij}$
to be 0 if nodes $i$ and $j$ are of different color.

As in other iterative methods, one has to choose the initial similarity scores $x^0_{ij}$.
In our method, we set $x^0_{ij}=1$ for all $i\in E_A$, $j\in E_B$. Though the choice
may seem arbitrary, note that in the first iteration it leads to intuitive results.
$$s^1_{in}=\frac{\min(id(i),id(j))}{\max(id(i),id(j))}\hspace{1cm}s^1_{out}=\frac{\min(od(i),od(j))}{\max(od(i),d(j))}$$
If, for instance, a node $i$ has 3 in-neighbors and a node $j$ has 5 in-neighbors, the in-similarity of nodes $i$ and $j$
in the first iteration will be $\frac{3}{5}$. We find that to be an intuitive choice if we don't yet know
anything about the similarities of the neighbor nodes --- in that case we can only reason about the number of neighbor
nodes.

The termination condition is $\max_{ij} |x^{k}_{ij}-x^{k-1}_{ij}| < \varepsilon$ for some chosen
precision $\varepsilon$. Alternative termination condition could be used too.

Note that our method has computationally more complex update rule compared to previous methods.
Other methods include summation of total $id(i)id(j)$ terms for in-neighbors and total $od(i)od(j)$ terms for out-neighbors.
In our method, we have to solve the assignment problem for $id(i)$ and $id(j)$ in-neighbors and for $od(i)$ and $od(j)$
out-neighbors. Since efficient algorithms for the assignment problem (mentioned in Section \ref{sec:relatedwork}) exist,
its complexity should not
be of big practical importance. Also, as it will be discussed in Section \ref{sec:evaluation}, for practical purposes,
in the case of dense graphs, one could switch to complement graphs (that are sparse in this case) and so reduce the
computation time.

\begin{example}
In order to illustrate our method, we applied it on example graphs (shown in Figure \ref{fig:zager}) used by Zager \cite{Zager08}.
The similarity scores for the nodes of the graphs are presented in Table \ref{tab:zager}.

\begin{figure}
\centering
\setlength{\unitlength}{0.7mm}
\begin{picture}(90,65)(10,0)
\put(10,40){\circle*{2}} \put(25,40){\circle*{2}}
\put(40,40){\circle*{2}} \put(10,40){\vector(1,0){15}}
\put(25,40){\vector(1,0){15}} \put(10,42){1} \put(25,42){2}
\put(40,42){3} \put(24,30){$G_A$}

\put(65,40){\circle*{2}} \put(65,42){1} \put(80,40){\circle*{2}}
\put(80,42){2} \put(110,40){\circle*{2}} \put(113,40){4}
\put(110,55){\circle*{2}} \put(113,55){3} \put(95,25){\circle*{2}}
\put(100,22){5} \put(95,10){\circle*{2}} \put(98,10){6}
\put(65,40){\vector(1,0){15}} \put(80,40){\vector(1,0){30}}
\put(80,40){\vector(1,-1){15}} \put(110,40){\vector(-1,-1){15}}
\put(110,55){\vector(0,-1){15}} \put(95,25){\vector(0,-1){15}}
\put(80,20){$G_B$}
\end{picture}
\caption{Two example graphs given by Zager \cite{Zager08}.}
\label{fig:zager}
\end{figure}
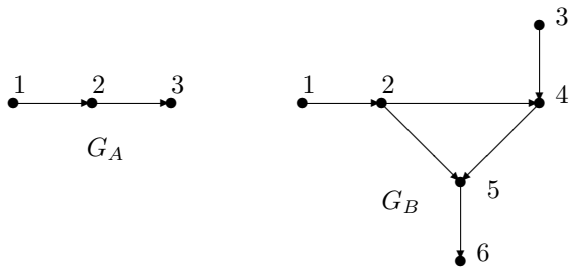

\begin{table}
\center
 \begin{tabular}{cccccccc}
  \hline
   & $1_B$ & $2_B$ & $3_B$ & $4_B$ & $5_B$ & $6_B$\\
  \hline
  $1_A$ & 0.682 & 0.100 & 0.597 & 0.200 & 0.000 & 0.000\\
  $2_A$ & 0.000 & 0.364 & 0.045 & 0.195 & 0.400 & 0.000\\
  $3_A$ & 0.000 & 0.000 & 0.000 & 0.091 & 0.091 & 0.700\\
  \hline
 \end{tabular}
\caption{Similarity scores for graphs given in Figure \ref{fig:zager}, calculated using
the method of neighbor matching for $\varepsilon=10^{-4}$.}
\label{tab:zager}
\end{table}
\end{example}

The proposed method converges, as stated by the following theorem.

\begin{theorem}
For any choice of graphs $G_A$ and $G_B$, for each pair of nodes $i\in V_A$ and $j\in V_B$, there exists
$x_{ij}=\lim_{k\rightarrow\infty}x^k_{ij}$ with a value in range $[0,1]$.
\label{thm:convergence}
\end{theorem}
\begin{proof}
For any $i\in V_A$ and $j\in V_B$, the corresponding sequence $(x^k_{ij})^\infty_{k=0}$ is nonincreasing.
We will prove this by induction on the number of iterations $k$.

The initial similarity score $x^0_{ij}$ for some $i$ and $j$ is equal to 1. The weight of the optimal matching when calculating
in or out similarity is equal $n_{in}$, or $n_{out}$ respectively, since the weight of the matching any two
nodes is 1. Since $m_{in}\geq n_{in}$ and $m_{out}\geq n_{out}$ it holds $s^{1}_{in}(i,j)=\frac{n_{in}}{m_{in}}\leq 1$
and $s^{1}_{out}(i,j)=\frac{n_{out}}{m_{out}}\leq 1$, and the same holds for $x^1_{ij}$ being the arithmetic mean
of the two values. This proves that in the first step, the similarity scores cannot grow, which is the base of the
induction.

Suppose that up to the step $k$ the sequence of scores $x^k_{ij}$ is nonincreasing, meaning that
$x^{k}_{ij}\leq x^{k-1}_{ij}$. This actually states that the weights of matching of any two nodes
when calculating $s^{k+1}_{in}$ and $s^{k+1}_{out}$ are not greater than the weights when calculating
$s^k_{in} $ and $s^k_{out}$, and thus $s^{k+1}_{in}\leq s^k_{in}$ and $s^{k+1}_{out}\leq s^k_{out}$.
We show this for in-similarity, and the reasoning for out-similarity is
analogous. Let $f^k_{ij}$ and $g^k_{ij}$ be the enumeration functions of the optimal matching of in-neighbors
of nodes $i\in V_A$ and $j\in V_B$ in iteration $k$. Then, it holds
$$\sum_{l=1}^{n_{in}}x^{k+1}_{f^{k+1}_{ij}(l)g^{k+1}_{ij}(l)}\leq\sum_{l=1}^{n_{in}}x^k_{f^{k+1}_{ij}(l)g^{k+1}_{ij}(l)}\leq\sum_{l=1}^{n_{in}}x^k_{f^k_{ij}(l)g^k_{ij}(l)}$$
The first inequality holds by inductive hypothesis, and the second by the optimality of the matching,
defined by $f^k_{ij}$ and $g^k_{ij}$, in iteration $k$.
Dividing all three expressions by $m_{in}$, we conclude $s^{k+1}_{in}(i,j)\leq s^k_{in}(i,j)$.
The same holds for out-similarities. Consequently, we have $x^{k+1}_{ij}\leq x^{k}_{ij}$.
This proves the inductive step. Hence, the sequence of similarity
scores $(x^{k}_{ij})^\infty_{k=0}$ is nonincreasing.

By induction on the number of iterations we prove that in all
the iterations, all the similarity scores are nonnegative. In the
first iteration, all the scores are nonnegative. In each subsequent
iteration, the update rule consists of averaging some of the scores
from the previous iteration. By averaging nonnegative values one cannot
obtain a negative value, so each sequence of similarity scores is nonnegative
and thus, bounded from below by zero. Nonincreasing sequence bounded from
below must have a limit, so $x_{ij}=\lim_{k\rightarrow\infty}x^k_{ij}$
exists. Since the sequence is nonincreasing and $x^0_{ij}=1$, the limit can't be greater than 1.
Also, since all the elements are nonnegative, the limit also has to be nonnegative. This proves
the theorem.
\end{proof}

Simple examples can be produced to show that the bounding interval $[0,1]$ is tight.

Important property of the similarity for isomorphic graphs is established by the following theorem.

\begin{theorem}
For two isomorphic graphs $G_A$ and $G_B$, let $f:V_A\rightarrow V_B$ be an isomorphism
between two graphs. For each node $i\in V_A$, it holds that $x_{if(i)}=1$.
\label{thm:isomorphism}
\end{theorem}
\begin{proof}
We show that $x^{k}_{if(i)}=1$ for all $i\in V_A$ and all $k\geq 0$ by induction on
the number of iterations $k$.

The initial value $x^0_{if(i)}$ is equal to 1 for all $i\in V_A$, by definition. This is the base
of the induction. Let $k>0$, assume $x^{k}_{if(i)}=1$ for all $i\in V_A$, and consider
$x^{k+1}_{if(i)}$. Since $f$ is an isomorphism of two graphs,
nodes $i$ and $f(i)$ must have the same number of in-neighbors and out-neighbors.
Hence, $m_{in}=n_{in}$ and $m_{out}=n_{out}$. It suffices to prove that the weights of the
optimal matchings when calculating in and out similarity are equal to $n_{in}$ and $n_{out}$
respectively. We discuss in-similarity first.
Since $f$ is the isomorphism, it maps all the in-neighbors of node $i$ to
in-neighbors of node $f(i)$. The weights $x_{af(a)}$ of matching each in-neighbor $a$ of $i$ to
in-neighbor $f(a)$ of $f(i)$ are equal to 1 by the inductive hypothesis, thus being maximal.
So the matching of each in-neighbor $a$ of $i$ to in-neighbor $f(a)$ of $f(i)$ is optimal.
Since there is $n_{in}$ in-neighbors, the weight of the optimal matching of in-neighbors
is $n_{in}$. Analogous reasoning is used to show that the weight of the optimal matching
of out-neighbors is equal to $n_{out}$. Therefore, both in and out similarity of $i$ and $f(i)$
in step $k+1$ are equal to 1 for all $i\in V_A$ and so, the similarity score $x^{k+1}_{if(i)}$ is also
equal to 1 for all $i\in V_A$.

Since $x^{k}_{if(i)}=1$ for all $k\geq 0$, and $i\in V_A$, the limit $x_{if(i)}$ is
also 1 for all $i\in V_A$.
\end{proof}

In the case $G_A=G_B$ where $f$ is the trivial automorphism $f(i)=i$ for all $i\in V_A$,
this theorem implies a simple corollary.

\begin{corollary}
For any graph $G_A$ and each node $i\in V_A$, it holds $x_{ii}=1$.
\end{corollary}

It is easy to check that the proven theorems hold for colored graphs too.

By the above statements, the neighbor matching method fulfills the first two requirements listed
in Section \ref{sec:relatedwork}. The matrix normalization is avoided and it is easy to produce examples 
of graphs with all the similarity values being 0 or all the similarity values being 1. Similarity
of nodes due to lack of in or out neighbors is recognized because in that case in or out
similarity will be equal to 1. So, we can conclude that all the requirements listed in Section
\ref{sec:relatedwork} are met.

\subsection{Measuring Similarity of Graphs}

The method of neighbor matching can be used to construct a similarity measure of two graphs
in the way of Heymans and Singh \cite{Heymans03}. When the similarity scores $x_{ij}$ for graphs $G_A$ and $G_B$
are computed, the optimal matching between their nodes can be found by solving the assignment problem
between the nodes from $V_A$ and $V_B$ with the weight of matching two nodes being the similarity of
the nodes. Let $f$ and $g$ be enumeration functions for the optimal matching and
$n=\min(|V_A|,|V_B|)$. Then, similarity of graphs $G_A$ and $G_B$ can be computed by
\begin{equation}
s(G_A,G_B)=\frac{1}{n}\sum_{l=1}^{n}x_{f(l)g(l)}.
\label{eq:grsim}
\end{equation}
By Theorem \ref{thm:convergence}, the value of the similarity measure $s$ is bounded in the
interval $[0,1]$. As a simple corollary of theorem \ref{thm:isomorphism}, if $G_A$ and
$G_B$ are isomorphic, it holds $s(G_A,G_B)=1$.

Of course, different similarity measures for graphs could be constructed based on
the similarities of their nodes. For instance, the sum of weights of the optimal matching
could be divided by $\max(|V_A|,|V_B|)$ instead of $\min(|V_A|,|V_B|)$. Such a choice
would penalize the difference in size when comparing two graphs. Another interesting
choice would be to take the average of all the values in the similarity matrix. In such
a case, graphs with greater number of automorphisms would be considered to be more
self-similar than graphs without automorphisms. In the rest of the paper we will use
the measure defined by the equation \ref{eq:grsim}.

\section{Experimental Evaluation}
\label{sec:evaluation}

We implemented the method of neighbor matching\footnote{The
source code of the implementation of the neighbor matching method is available from
\url{http://www.matf.bg.ac.rs/~nikolic/software.html}.}
and the methods of Zager and Verghese and of Heymans and Singh in C++.\footnote{
The C++ implementation of the method of Heymans and Singh was
obtained by a simple transformation of Java implementation kindly 
provided by Ambuj Singh.}
For solving the assignment problem, we used an available implementation of the 
Hungarian algorithm \cite{Knuth93}. Nevertheless, more efficient algorithms 
(mentioned in Section \ref{sec:preliminaries}) exist.

In this section, we describe two experiments we performed to test
the performance of our method. The first one was related to the matching of the
isomorphic subgraph, and the second one was the classification of the
Boolean formulae.

\subsection{Isomorphic Subgraph Matching}

Here we present a slightly modified experiment from Zager and Verghese \cite{Zager08}
which we use to compare several methods for computing node similarity.
We will consider a problem of finding a subgraph of a graph $A$ that is isomorphic
to some other graph $B$. We will use random Erd\H{o}s--R\'enyi graphs $G_{n,p}$.
The experiment consists of generating
a random graph $A$ of size $n$ and randomly selecting $m\leq n$ nodes which
induce a subgraph $B$ of $A$. The similarity of nodes of $A$ and $B$ is calculated,
the assignment problem between the nodes of $A$ and $B$ is solved, and the
matching of the nodes is obtained. Then, it is checked if graph $B$ is isomorphic
to the subgraph of $A$ induced by the obtained matching.

For $n=15$, this procedure is repeated 500 times for each pair of
$m=8,9,\ldots,15$ and $p=0.2,0.4,0.6,0.8$, and the accuracy of the
method (the percentage of correct guesses) is calculated for each pair.
Required numeric precision when calculating similarities for all the methods was
$\varepsilon=10^{-4}$, and the same termination condition was used ---
$\max_{ij} |x^{k}_{ij}-x^{k-1}_{ij}| < \varepsilon$.

The methods compared were the method of neighbor matching (NM), the one of Heymans and Singh
(HS), and the one of Zager and Verghese (ZV). It was noted that NM and ZV methods
are heavily influenced by density parameter $p$ both
in matching performance and speed, while the HS method is not.
We believed that it is due to the fact that HS method is considering
both the input graphs and their complements. As suggested in Section
\ref{sec:method}, we made a modification to other two methods which we call
``the complement trick'' --- for dense graphs ($p>0.5$) the similarity of nodes is
measured for the complement graphs instead of the original input graphs.\footnote{
The complement trick could be given an intuitive rationale. For
instance, consider one trying to reason about similarity of two sparse
graphs based on their adjacency matrices. Probably, one would spot ones
in the matrices and analyze their arrangements in some way. If the graphs
were dense it would be much easier to spot zeroes and reason about them.}
This introduced methods NM* and ZV*. For completeness of the evaluation,
we introduced HS*, too.

\begin{figure}
 \includegraphics[width=0.48\textwidth]{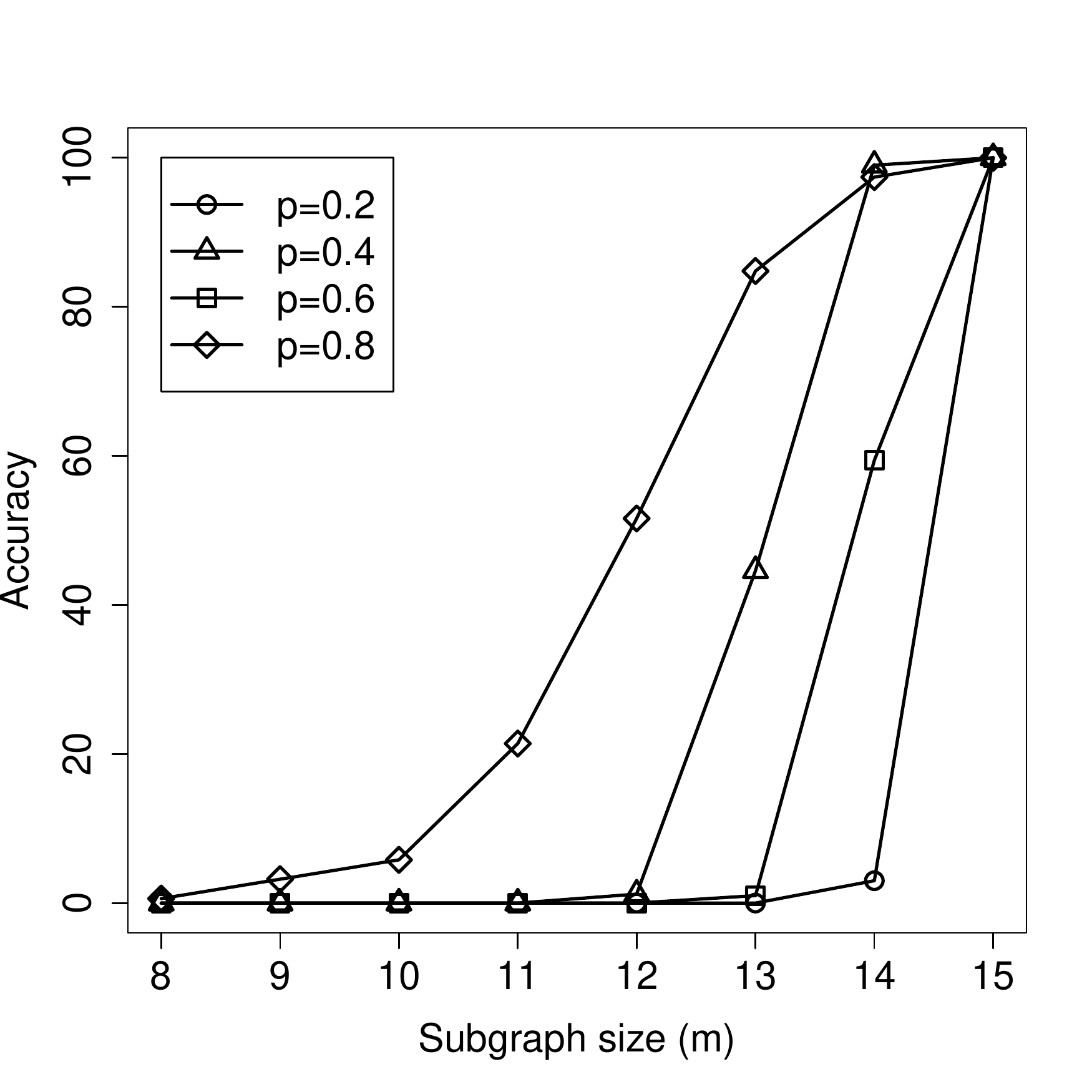}
 \includegraphics[width=0.48\textwidth]{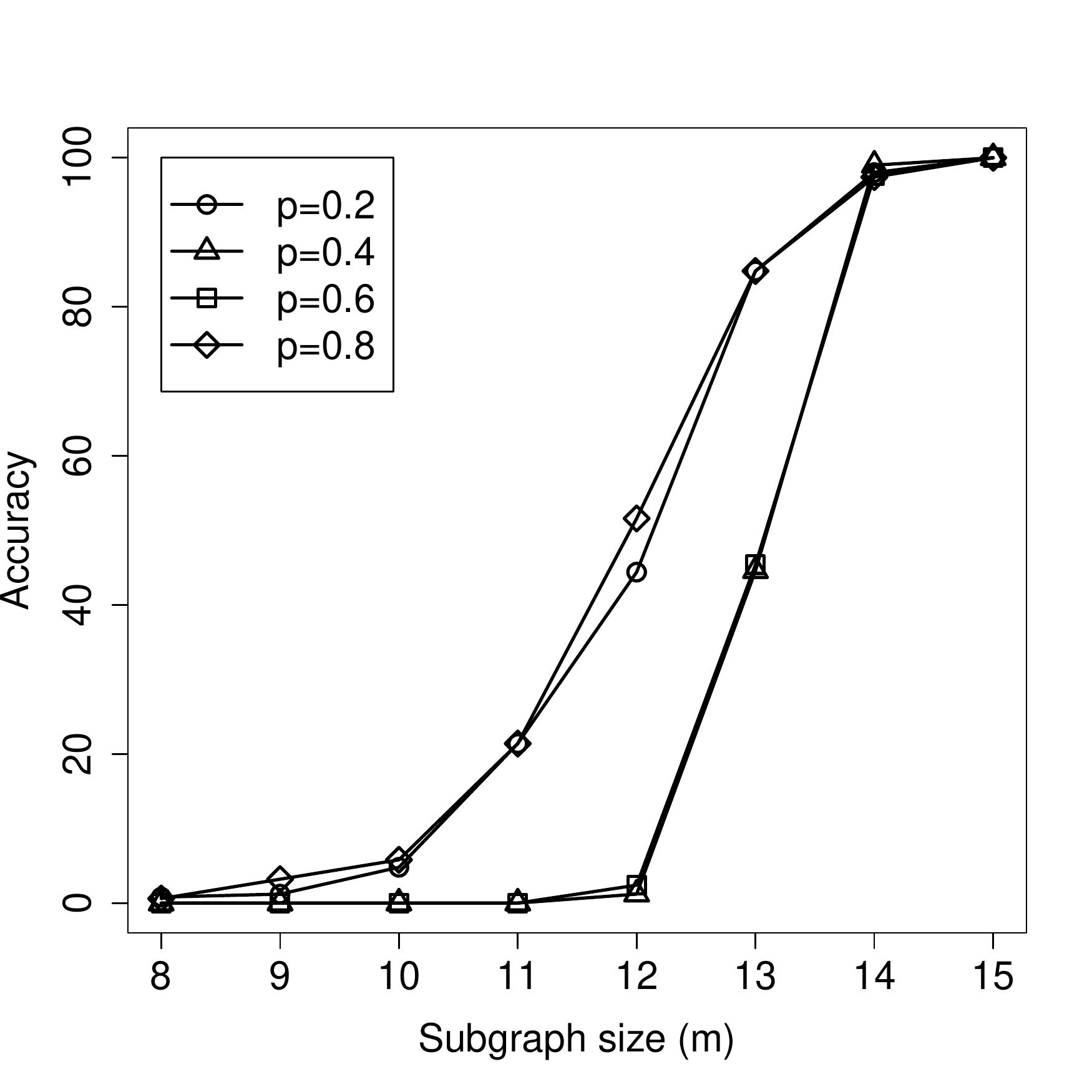}
\caption{Accuracy of isomorphic subgraph matching for NM and NM* methods.}
\label{fig:nm}
 \includegraphics[width=0.48\textwidth]{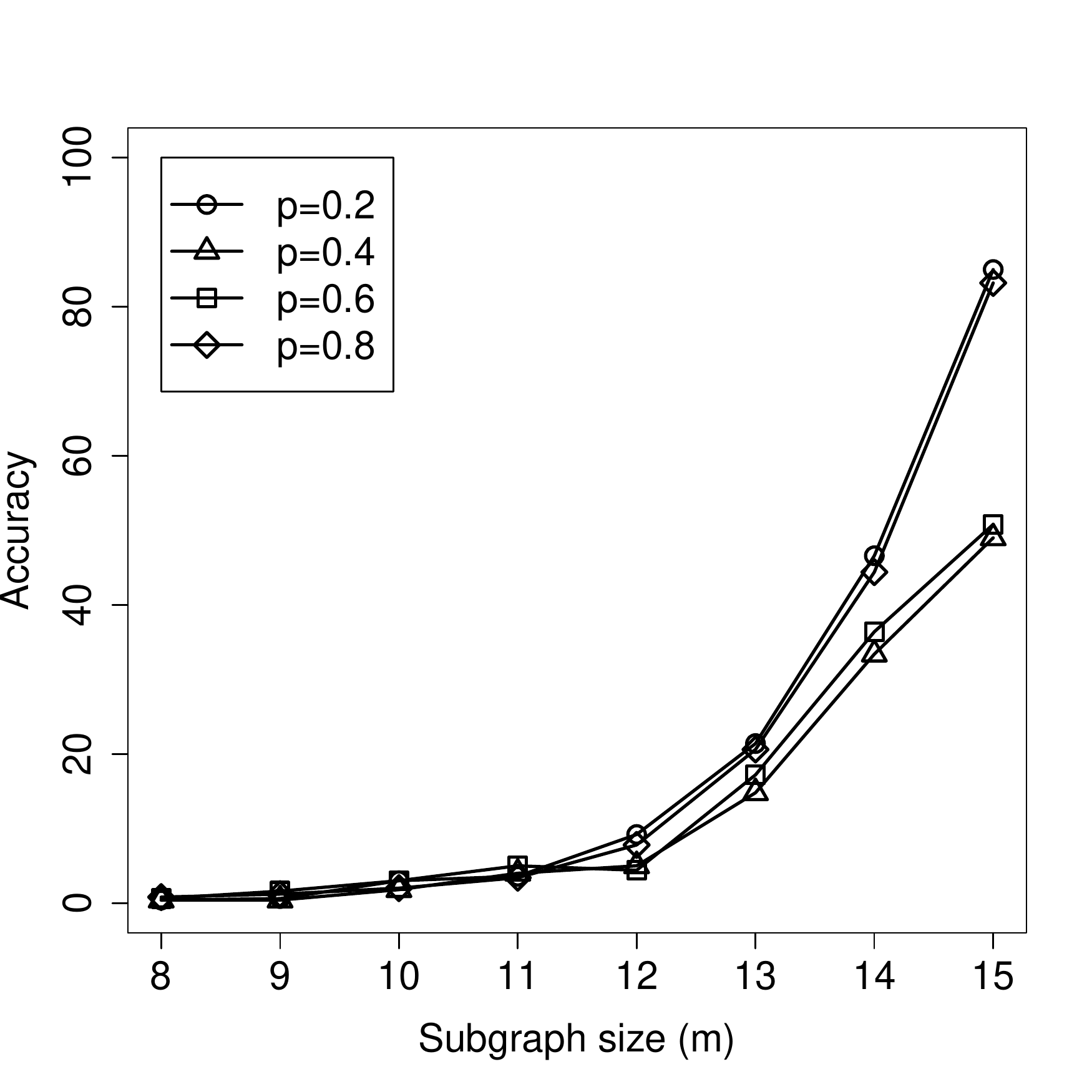}
 \includegraphics[width=0.48\textwidth]{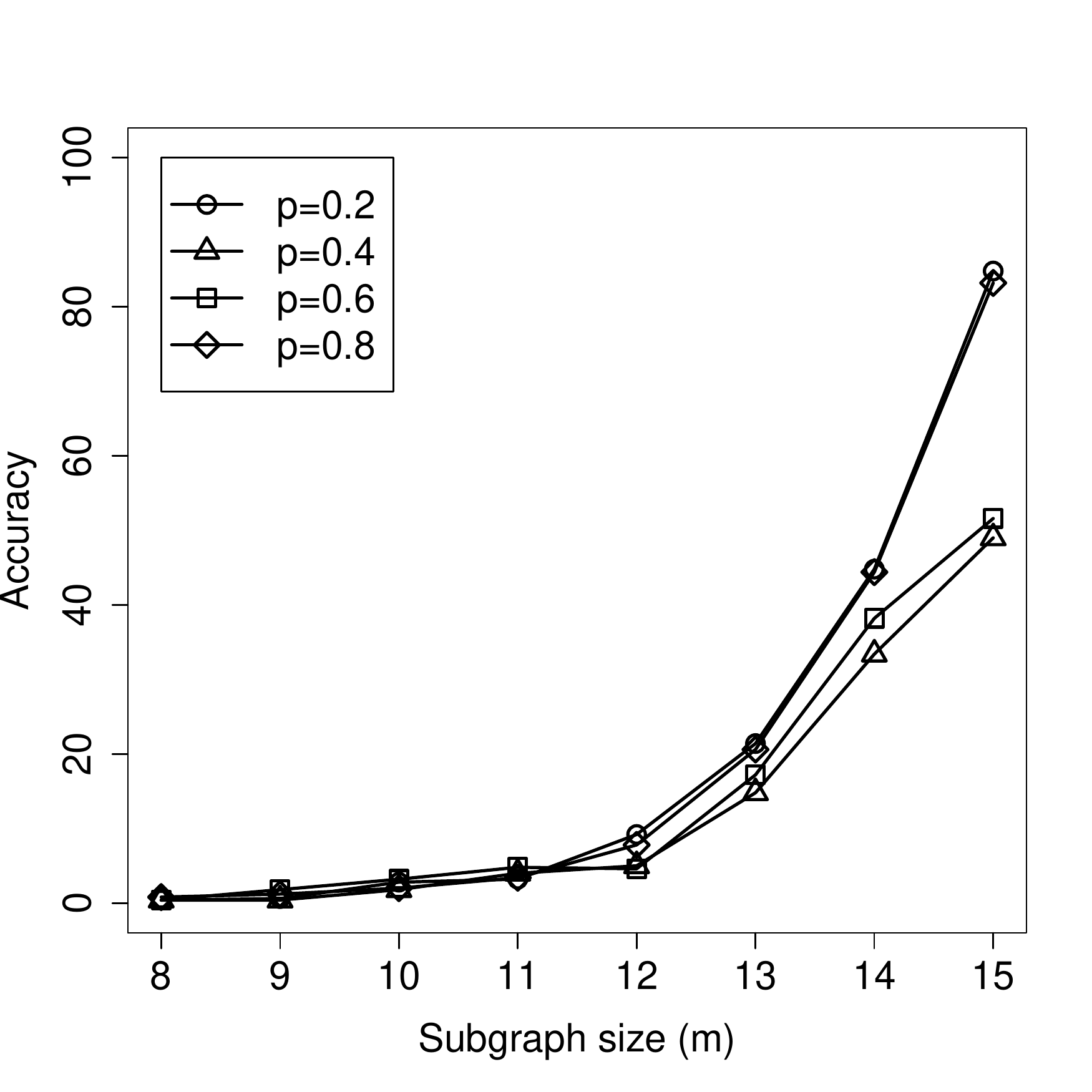}
\caption{Accuracy of isomorphic subgraph matching for HS and HS* methods.}
\label{fig:hs}
 \includegraphics[width=0.48\textwidth]{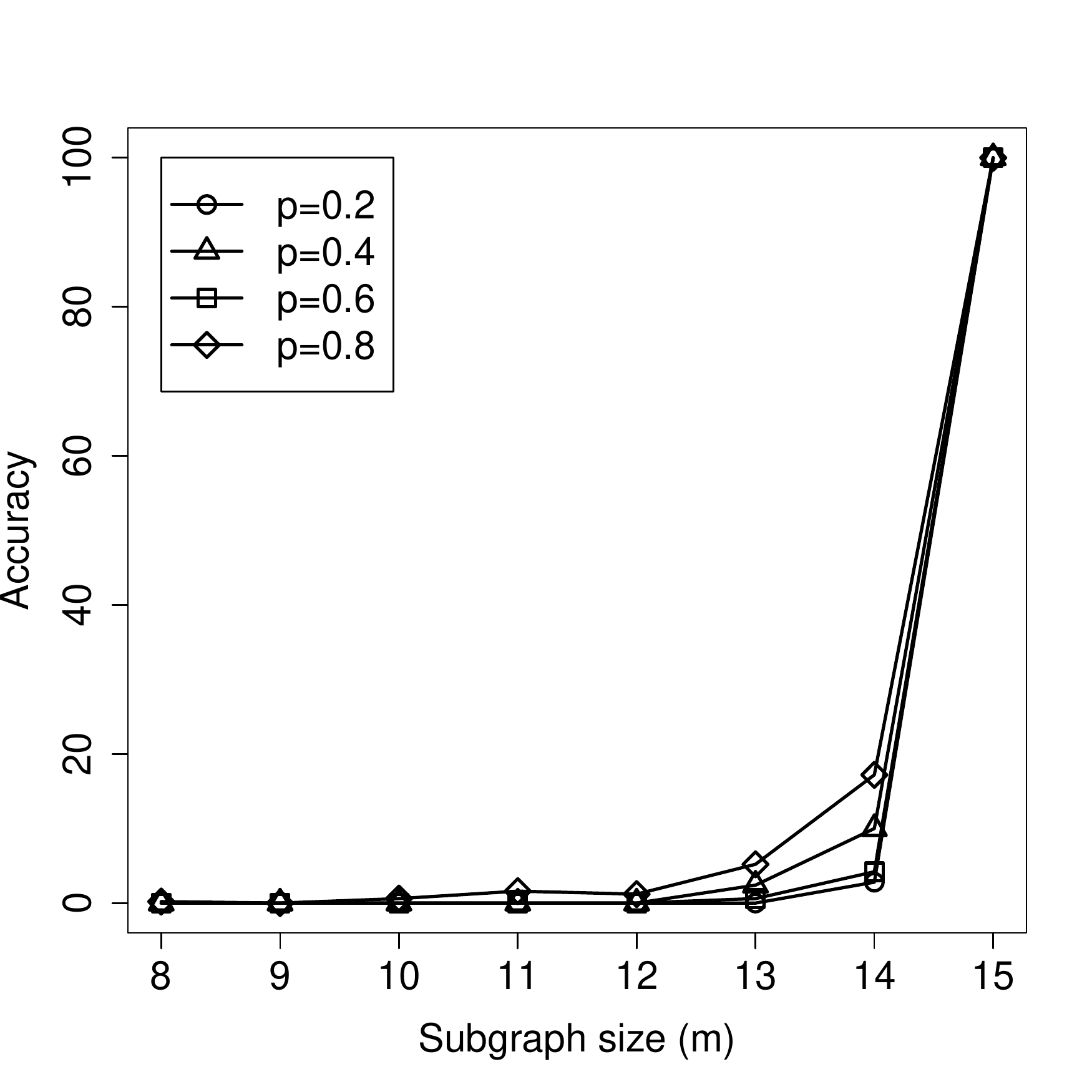}
 \includegraphics[width=0.48\textwidth]{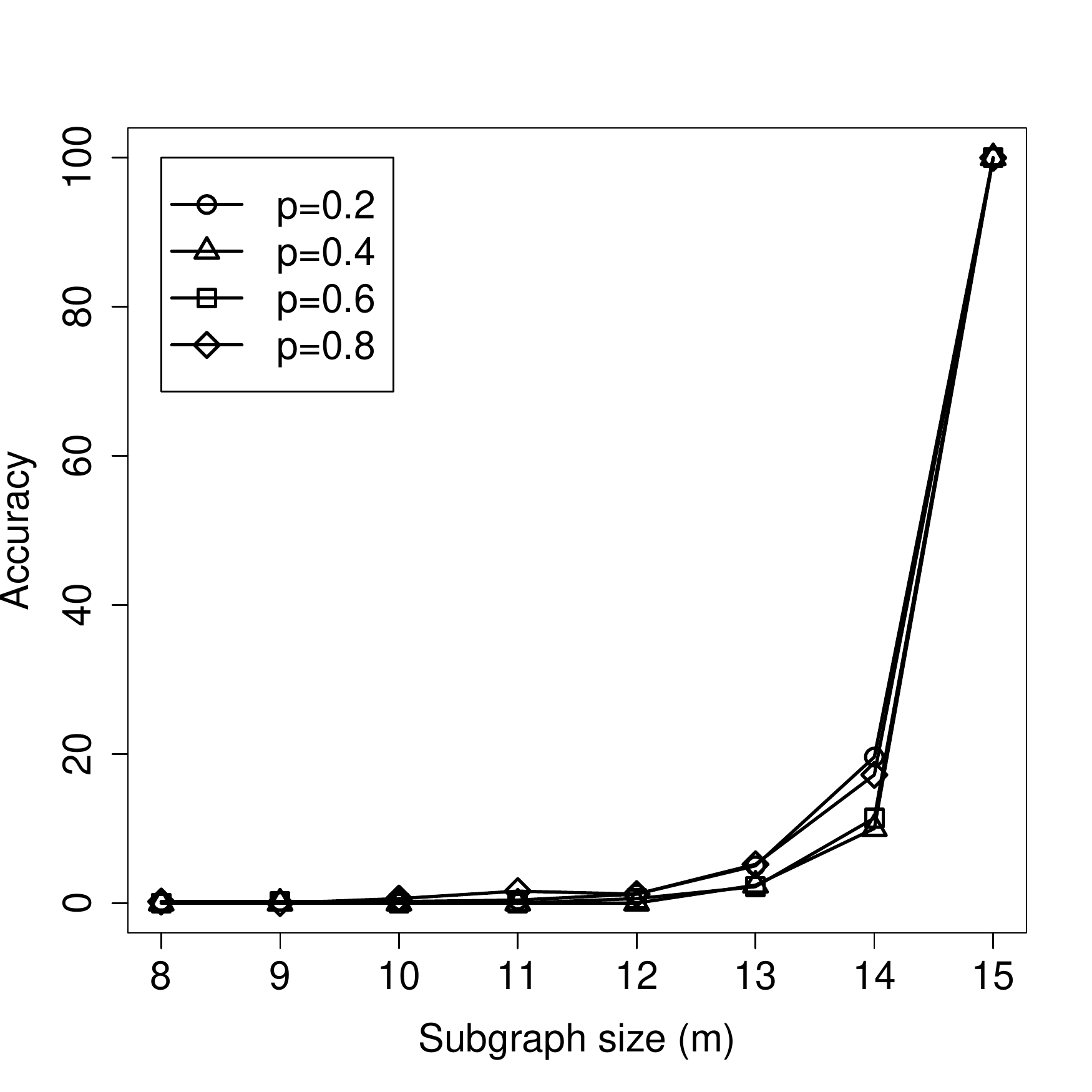}
\caption{Accuracy of isomorphic subgraph matching for ZV and ZV* methods.}
\label{fig:zg}
\end{figure}

For each method, for each value of parameter $p$, we present one plot that
shows the percentage of successes in isomorphic subgraph matching for each value
of $m$. The plots are presented in figures \ref{fig:nm},\ref{fig:hs}, and
\ref{fig:zg}. It can be noted that the accuracy of ZV and ZV* generally rises
much later than for the other methods. NM* obviously performs the best.

In Table \ref{tab:results}, for each method, we present the overall accuracy in the
experiment and the total time spent for the experiment.

\begin{table}[h!]
\centering
 \begin{tabular}{lcccccc}
  \hline
  & NM & NM* & HS & HS* & ZV & ZV*\\
  \hline
  Accuracy & 27.3 & 37.8 & 17.5 & 17.5 & 13.9 & 15.0\\
  Time & 2062s & 838s & 11511s & 11730s & 349s & 230s \\
  \hline
 \end{tabular}
\caption{Overall accuracy and time needed for the experiment, for each method used.}
\label{tab:results}
\end{table}

The complement trick obviously improved NM and ZV methods. As expected,
it did not affect the HS method. For NM*
and ZV* methods, apart from boosting the accuracy, the computation time is significantly reduced. For NM method,
this modification reduces the computation time for solving the assignment
problem in NM update rule, since it reduces the number of nodes to be matched
in the cases when this number can be large (dense graphs).

\subsection{The Classification of Boolean Formulae}

Here we present the problem of the classification of Boolean formulae which we
use to show that our method can capture a meaningful similarity in a real world
problem.

Various important practical problems can be modeled in Boolean logic
including problems in electronic design automation, software and hardware
verification, scheduling, timetabling, artificial intelligence,
and other domains. Each instance of the problem is represented by
a Boolean formula. Classification of Boolean formulae
has been investigated in order to automatically tune SAT solvers
(systems for checking the satisfiability of Boolean formulae) that
is a practically important and challenging problem. A very reliable
approach to Boolean formulae classification is based on measuring the distances
between the formulae \cite{Nikolic09}. In that approach, in order to compute 
the distance between the formulae, they are represented by numerical vectors
of some syntactical features, that can be computed for each formula. However,
Boolean formulae have a natural variable-clause graph representation \cite{Nudelman04}
that could be used for their classification.

We performed the classification of Boolean formulae using our similarity
measure for graphs on their graph representation. We used 149 structured instances from
SAT competition 2002 benchmark set (which is one of the standard benchmarks sets
for SAT).\footnote{The benchmarks are available from \url{http://www.satcompetition.org}.}
Most of the formulae had up to 1000 nodes, but 25 of them were larger (up to 5280 nodes).
Formulae were grouped in 9 classes corresponding to the problems the formulae
originate from. Graphs corresponding to the formulae had from 122 to 5280 nodes.
Differences in graph size of order of magnitude were present within each class
too. The classification was performed using the $k$ nearest neighbors algorithm
with leave one out evaluation procedure ---
for each formula $F$, its graph similarity to the remaining formulae was computed,
and the set $N(k)$ of $k$ most similar formulae was determined.
Formula $F$ is classified to the class that has the most representatives in the set $N(k)$.
For the evaluation of the classification performance, we measured
the accuracy of the classification --- number of correctly classified formulae divided by the
total number of formulae being classified.

The best accuracy of the classification was 93\% for $k=7$.
The best accuracy for a domain specific approach from \cite{Nikolic09} on the same set
is 96\% for $k=1$. Only slightly more accurate, the domain specific approach is based on long lasting
research in the field \cite{Nudelman04,Xu08,Nikolic09}. It is interesting
to see that the general approach, not designed specifically for this purpose,
can achieve a very high accuracy.
Most importantly, we confirmed that our similarity measure can capture
a meaningful similarity in a real world problem.

A very interesting remark concerning this experiment is that the difference
in size of the compared graphs did not influence the adequateness of the
similarity measure. This kind of robustness might be interesting for practical
applications.

\section{Conclusions and Future Work}
\label{sec:conclusions}

We proposed a refined notion of similarity of graph nodes, and
based on that refinement we developed a new iterative method for measuring similarity
of nodes of two graphs. This method was extended to a method for measuring similarity
of whole graphs. We proved the convergence of the method and showed that it has several
desirable properties (listed in Section \ref{sec:relatedwork}) that, to our knowledge,
the existing methods lack.

We implemented the method and evaluated the implementation on two test problems.
On one test problem (the isomorphic subgraph matching problem), we confirmed that
the proposed method performs better than other methods. On the second one, it is
confirmed that the graph similarity measure is able to capture a meaningful
similarity in a real world problem. The method showed to be robust to differences
in graph size. The performance on dense
graphs can be significantly boosted by measuring the similarity of nodes of complement
graphs. This modification can significantly reduce the running time of the method.

As for the future work, we are planning applications of the neighbor matching method
in real-world problems in bioinformatics, text classification, and other domains
suitable for graph similarity techniques.

\section*{Acknowledgements}
This work was partially supported by Serbian Ministry of Science grant 144030.

\bibliographystyle{elsarticle-num}
\bibliography{similarity}

\end{document}